\documentclass[10pt,twocolumn,letterpaper]{article}
\usepackage{capt-of}
\usepackage{placeins}
\usepackage{cvpr}              
\usepackage[accsupp]{axessibility}

\usepackage{float}
\usepackage[T1]{fontenc}

\definecolor{cvprblue}{rgb}{0.21,0.49,0.74}
\usepackage[pagebackref,breaklinks,colorlinks,allcolors=cvprblue]{hyperref}

\def\paperID{MIV-12} 
\def\confName{CVPR}
\def\confYear{2025}

\title{Decoding Vision Transformers: the Diffusion Steering Lens}

\author{
Ryota Takatsuki\textsuperscript{1,2,3}\thanks{Corresponding author: \texttt{takatsuki\_ryota@araya.org}} \qquad Sonia Joseph\textsuperscript{4,5} \qquad Ippei Fujisawa\textsuperscript{1,2} \qquad Ryota Kanai\textsuperscript{1,2} \\
[0.8em]
\textsuperscript{1}Araya Inc. \qquad
\textsuperscript{2}AI Alignment Network \qquad
\textsuperscript{3}The University of Tokyo \\
\textsuperscript{4}Mila - Quebec AI Institute \qquad
\textsuperscript{5}McGill University
}

\begin{document}
\maketitle
\begin{abstract}
Logit Lens is a widely adopted method for mechanistic interpretability of transformer-based language models, enabling the analysis of how internal representations evolve across layers by projecting them into the output vocabulary space. Although applying Logit Lens to Vision Transformers (ViTs) is technically straightforward, its direct use faces limitations in capturing the richness of visual representations. Building on the work of Toker et al. (2024)~\cite{Toker2024-ve}, who introduced Diffusion Lens to visualize intermediate representations in the text encoders of text-to-image diffusion models, we demonstrate that while Diffusion Lens can effectively visualize residual stream representations in image encoders, it fails to capture the direct contributions of individual submodules. To overcome this limitation, we propose \textbf{Diffusion Steering Lens} (DSL), a novel, training-free approach that steers submodule outputs and patches subsequent indirect contributions. We validate our method through interventional studies, showing that DSL provides an intuitive and reliable interpretation of the internal processing in ViTs.
\footnote{Our code is available at https://github.com/rtakatsky/DSLens.}
\end{abstract}
\vspace{-2mm}

\section{Introduction\label{sec:intro}}

\begin{figure}
    \centering
    \includegraphics[width=0.8\linewidth]{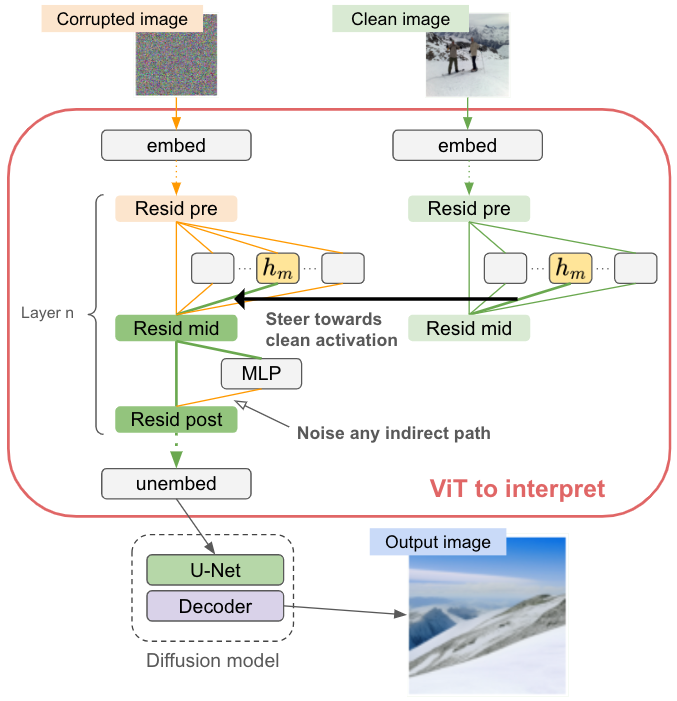}
    \caption{\textit{Schematic of Diffusion Steering Lens.} Visualizing the direct contribution of \(h_m\) at layer \(n\).}
    \label{fig:dsl}
    \vspace{-5mm}
\end{figure}

Mechanistic interpretability has emerged as a crucial framework for understanding the internal computations of deep neural networks~\cite{bereska2024mechanisticinterpretabilityaisafety}. A widely used technique in the domain is Logit Lens~\cite{nostalgebraist2020-cu}, which projects intermediate representations of transformer-based models into the output vocabulary space. This method is grounded in the insight that the residual stream in transformer architectures performs an iterative refinement of predictions~\cite{Belrose2023-ow,Lad2024-fe,Rushing2024-tg}—a perspective supported by layer ablation and swapping studies~\cite{Zhao2021-wx,Lad2024-fe,Sun2024-ni} as well as observations in other residual networks~\cite{Greff2016-bz,Jastrzebski2017-mh}. Tracking these internal "belief" updates provides an intuitive window into how models process information.

Although the internal processing of deep convolutional neural networks has been extensively investigated~\cite{Cammarata2020-jc,Cammarata2021-em,Gorton2024-fo,Schubert2021-hc,Rajaram2024-gl,Bau2017-lo}, the internal mechanisms of Vision Transformers (ViTs) remain relatively underexplored. While Logit Lens has been applied to ViTs~\cite{Joseph2024-no}, its direct application is insufficient to capture the full richness of visual processing—visual representations span a feature space far more complex than classification labels alone. This issue can be mitigated by training numerous probes, but it remains unclear which features—and how many—should be employed. In contrast, visually decoding the internal representations can offer a more comprehensive interpretation of the complex visual information processed in ViTs.

In this study, we propose a novel, training-free technique for interpreting the iterative refinement of predictions in ViTs. Our approach builds on "Diffusion Lens"~\cite{Toker2024-ve}, a method originally proposed for interpreting text encoders in text-to-image (T2I) diffusion models. We first demonstrate that Diffusion Lens effectively visualizes residual stream representations in ViTs, yet it fails to provide interpretable visualizations for the direct contributions of individual submodules such as attention heads~\cite{Elhage2021-py}. To overcome this shortcoming, we introduce \textbf{Diffusion Steering Lens} (DSL), a method that steers submodule outputs and patches subsequent contributions to isolate their direct effects, thereby providing a more precise and interpretable view of how ViTs build their final predictions. We validate DSL through rigorous interventional studies, demonstrating that it reliably highlights the contributions of attention heads.

\section{Related Work}
\subsection{Inspection Methods of Transformers}
Mechanistic interpretability has leveraged various methods to inspect intermediate activations of transformer models. A common approach to date involves training linear probes on internal representations~\cite{Alain2016-nd,Belinkov2021-bs}. In transformer interpretability, projecting intermediate representations into the output vocabulary space—known as Logit Lens~\cite{nostalgebraist2020-cu}—has proven particularly effective. Recent work has proposed improvements that mitigate representation drift via linear transformations~\cite{Belrose2023-ow} and enhance expressivity by patching representations across contexts, layers, and models~\cite{Ghandeharioun2024-ei}. 

\subsection{Interpretation Through Output}

Logit lens exemplifies the shift from input-based to output-centric interpretability. Early work in mechanistic interpretability of vision models often employed activation maximization~\cite{Olah2017-wd,Nguyen2019-bs} to identify patterns that maximally activate specific units. Despite their success, such feature visualization methods have been criticized for not always reflecting the actual internal processing~\cite{Zimmermann2021-ub,Geirhos2023-ys}. Another class of input-based interpretability relies on heatmaps in the input image to highlight attributions to specific representations~\cite{Ribeiro2016-ax,Lundberg2017-lo,Selvaraju2016-px,Chefer2020-rr}. However, input attribution alone does not fully reveal how specific information is transformed by internal computations or how it contributes to the final output. These limitations have motivated output-centric interpretability techniques, where the roles of internal components are inferred by analyzing how modifying their activations affects the final output.~\cite{Gur-Arieh2025-yb}. This is commonly done by steering~\cite{Turner2023-od}, which involves altering activations along specific directions in representation space. Yet, since final predictions result from a combination of direct and indirect effects, isolating the direct contribution of a specific submodule requires further methodological care. Another output-centric approach that is relevant but distinct from our approach is text-based decomposition methods~\cite{Gandelsman2023-gt, Gandelsman2024-rz}, which propose to describe CLIP sub-components' roles in text.

\subsection{Interpretation using diffusion model}

Toker et al.~\cite{Toker2024-ve} introduced Diffusion Lens as a means to interpret the text encoder in text-to-image (T2I) models. Their method applies the final layer normalization to the residual stream of a T2I encoder and then decodes these representations to illustrate the evolution across layers. They did not apply this method to visualize direct contributions from submodules, which we apply to an image encoder in this study. In related work, Daujotas~\cite{Daujotas2024-sa} explored the visualization of sparse autoencoder features in the image encoder of a multimodal diffusion model by steering a specific feature and decoding the output with a diffusion-based decoder. While training sparse autoencoders on intermediate representations of ViTs facilitates identifying fine-grained, interpretable units, as shown in the context of language model interpretability~\cite{Cunningham2023-ia,bricken2023monosemanticity,gao2024scalingevaluatingsparseautoencoders,templeton2024scaling}, our objective is to develop a tool that directly tracks how ViTs refine their predictions layer by layer.


\section{Method\label{sec:method}}
Our approach assumes the image encoder of the standard ViT architecture~\cite{Dosovitskiy2020-xn} in a multimodal diffusion model. Each transformer block processes the input residual stream \(\mathbf{x}^l_{\text{pre}}\) via self-attention heads and an MLP. Specifically, the block computes intermediate representations:
\begin{equation}
\mathbf{x}^l_{\text{mid}} = \mathbf{x}^l_{\text{pre}} + \sum_{h \in H_l} h(\mathbf{x}^l_{\text{pre}}), \quad
\mathbf{x}^l_{\text{post}} = \mathbf{x}^l_{\text{mid}} + m^l(\mathbf{x}^l_{\text{mid}}),
\end{equation}
where $\mathbf{x}^l_{\text{post}}$ becomes the input to the next layer. Our goal is to interpret the contributions of specific submodules—such as attention heads or MLPs—toward the final prediction along with the iterative refinement on the residual stream.
\subsection{Diffusion Steering Lens}
Figure~\ref{fig:dsl} illustrates the schematic of Diffusion Steering Lens (DSL). 
Let \(i\) denote a target (or clean) image and \(i^*\) a corrupted version (\(i^*\) is sampled from a Gaussian distribution in this study). For an internal representation \(x\) and its corrupted counterpart \(x^*\) we steer \(x^*\) toward \(x\):
\begin{equation}
x_{\text{new}} \leftarrow x^* + \alpha \cdot (x - x^*)
\end{equation}
where \(\alpha\) is a steering coefficient (set to 100 in our experiments). We find that low \(\alpha\) leads to noisy outputs, whereas larger values \(\alpha>10\) produce stable results.

To isolate direct submodule contributions, we replace any subsequent output \(o\) from an attention head or MLP following \(x\) with its corrupted counterpart \(o^*\).
After the steered representation \(x_{\text{new}}\) is propagated through the model, the diffusion-based decoder translates the resulting output into an interpretable image (In all our experiments, we consistently use 25 inference steps during the diffusion process).


\section{Visualization of Direct Contributions\label{sec:visualization}}
In this section, we qualitatively compare Diffusion Lens~\cite{Toker2024-ve} and our proposed Diffusion Steering Lens (DSL) on a Vision Transformer model. Throughout the paper, we apply both methods to CLIP ViT-bigG/14~\cite{Radford2021LearningTV}, the image encoder of Kandinsky2.2~\cite{Razzhigaev2023-ou} consisting of 48 transformer blocks and 16 attention heads in each block. We use 8 samples from the ImageNet Validation dataset~\cite{Deng2009-tb}.

\subsection{Diffusion Lens}
\begin{figure}
    \vspace{-3mm}    
    \centering
    \includegraphics[width=0.75\linewidth]{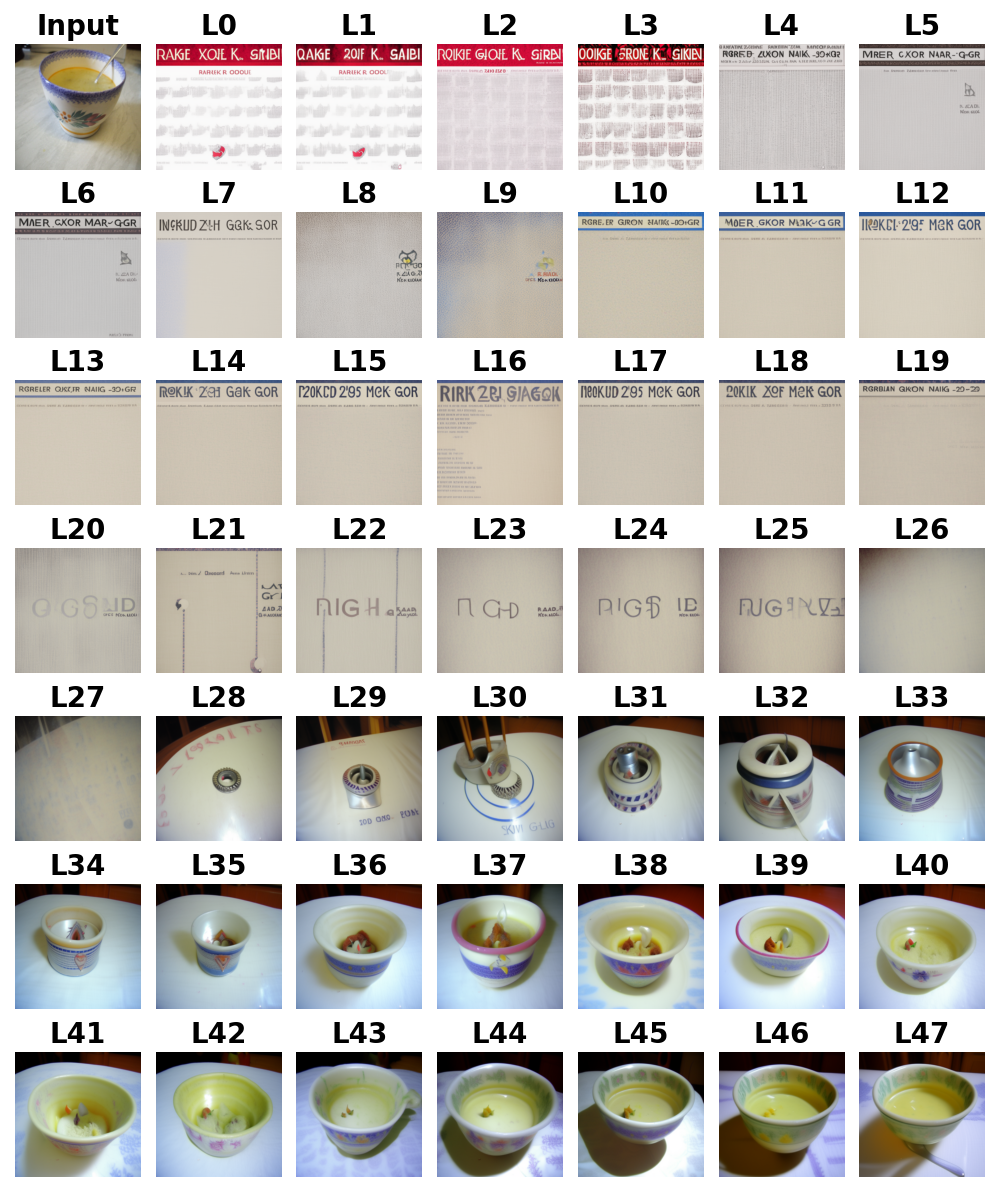}
    \caption{\textit{Diffusion Lens on Resid post: an example (cup)}. Around layer 30, the visualization starts resembling the input image.}
    \label{fig:diffusion_lens_resid_post}
    \vspace{-1mm}
\end{figure}

Figure~\ref{fig:diffusion_lens_resid_post} presents an example visualization of the residual stream representations (See Supplementary for other examples). Notably, around layer 30, the Resid post representation begins to resemble the input image. This observation contrasts with those in text encoders~\cite{Toker2024-ve}, where Diffusion Lens yields meaningful visualizations from layer 0. This difference is likely attributable to the distinct nature of visual versus textual modalities.

\begin{figure}
    \centering
    \includegraphics[width=1.00\linewidth]{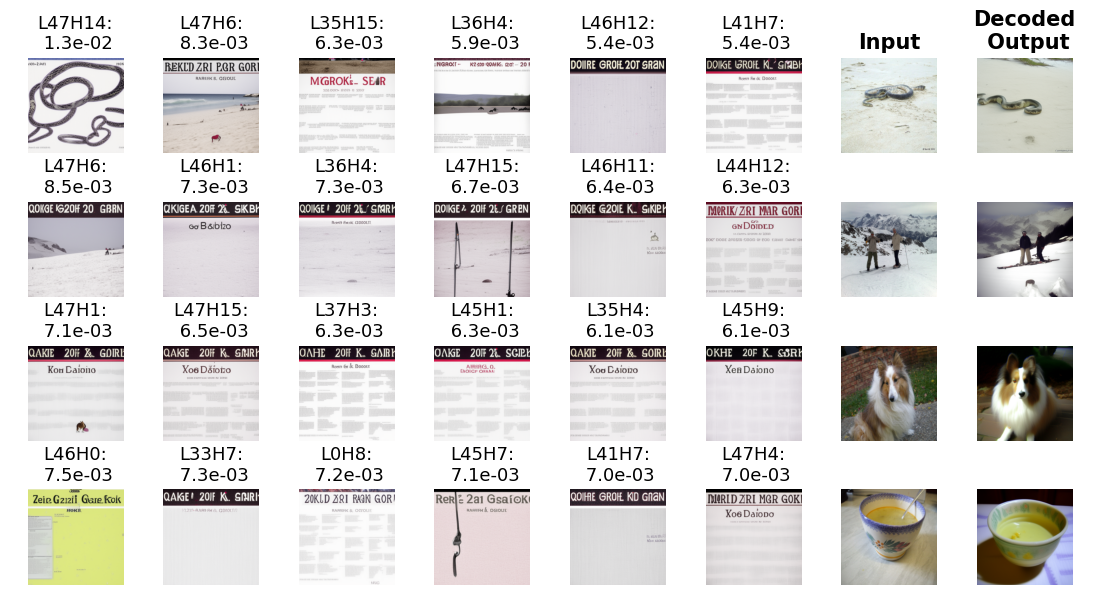}
    \caption{\textit{Top 6 heads in similarity with input when using Diffusion Lens}. Although a few heads in later layers reflect some contribution (e.g., L47H14: "snake"), most submodule outputs do not match the residual stream visualization.}
    \label{fig:top_6_heads_dl}
    \vspace{-2mm}
\end{figure}

Figure~\ref{fig:top_6_heads_dl} shows the 6 head visualizations most similar to the input image, as measured by cosine similarity in the output space of CLIP ViT-bigG/14 (See Supplementary for additional attention head and MLP visualizations). Although some heads in the final layers (e.g., L47H14, which produces a "snake"-like visualization) demonstrate meaningful contributions, many submodules yield similar outputs that diverge from the input image. Moreover, these visualizations do not consistently align with the residual stream representations, questioning their reliability.

\subsection{Diffusion Steering Lens}
Figure~\ref{fig:top_6_heads_dsl} shows the 6 head visualizations most similar to the input image when using DSL (See Supplementary for additional attention head and MLP visualizations). This shows that DSL facilitates identifying heads that have important contributions, that Diffusion Lens misses. Notably, contributions from earlier heads (e.g., L36H4) more closely align with the Resid post visualization. This comparison suggests that DSL offers a more reliable interpretation of how individual submodules contribute to the final prediction.

\begin{figure}
    \centering
    \includegraphics[width=1.00\linewidth]{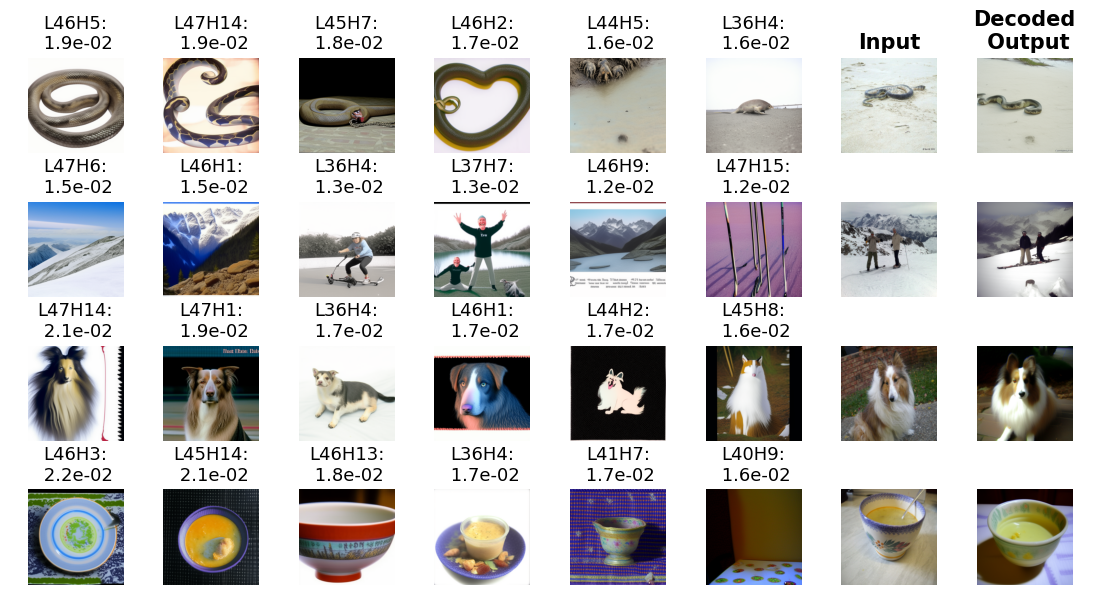}
    \caption{\textit{Top 6 heads in similarity with input when using Diffusion Steering Lens}. The DSL outputs more faithfully reflect the input, particularly for earlier heads (e.g., L36H4), highlighting its ability to visualize direct submodule contributions.}
    \label{fig:top_6_heads_dsl}
    \vspace{-2mm}
\end{figure}

\section{Evaluations\label{sec:eval}}
\subsection{Evaluation methods\label{sec:eval_methods}}
We evaluate the reliability of our Diffusion Steering Lens (DSL) using two interventional experiments that focus on the attention heads’ direct contributions.

\subsubsection{Evaluation 1} 
We test the hypothesis that ablating the output of attention heads whose visualizations closely resemble the input image will affect the final output more than ablating those with dissimilar visualizations. To quantify this effect, we measure the cosine similarity between the original output and the output after ablation in the CLIP model’s embedding space and see its correlation with the cosine similarity between the input image and the head visualization. We also conduct this evaluation on Diffusion Lens for comparison.

\subsubsection{Evaluation 2} 
\begin{figure}
    \centering
    \includegraphics[width=1.00\linewidth]{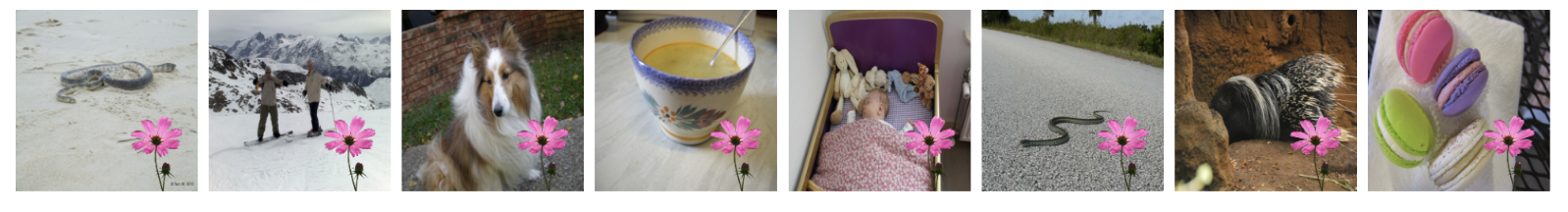}
    \caption{\textit{Images with overlays (flower)}}
    \label{fig:imagenet_with_overlays}
    \vspace{-4mm}
\end{figure}

We further assess the specificity of the DSL visualizations by introducing a synthetic overlay on ImageNet samples (Figure~\ref{fig:imagenet_with_overlays}). We sequentially ablate the outputs of attention heads in descending order—first by layer (from the highest to the lowest) and then within each layer by head index (from highest to lowest). We hypothesize that these progressive interventions will gradually remove the overlay’s influence, restoring the output to more closely resemble the original ImageNet image. We employ LLaVA-NeXT~\cite{liu2024llavanext} to automatically identify the attention heads whose visualizations capture the synthetic overlay. Given each head visualization, we ablate heads that receive a “yes” response from the following VQA question: “Is the main subject of this image a clearly visible \texttt{OVERLAY\_NAME}? Ignore small, blurry, or background ones. Answer only 'yes' or 'no'.” (e.g. \texttt{OVERLAY\_NAME} = "flower"). We compare DSL against the following two baselines: 
\begin{itemize}
\item \textbf{ACDC~\cite{conmy2023automatedcircuitdiscoverymechanistic}-like Ablation:} We set a threshold \(\tau\) on the loss of cosine similarity with the source output and select, for ablation, the head whose removal results in an increase in cosine similarity with the target output within this threshold. This presents a theoretically optimal case.
\item \textbf{Random Ablation:} For each layer, we randomly select the same number of heads as in the DSL case, but only from those heads that received a “no” response to the VQA query. This selection is repeated 50 times to obtain a robust baseline.
\end{itemize}

\subsection{Results\label{sec:results}}
\subsubsection{Evaluation 1} 
\begin{figure}
    \centering
    \includegraphics[width=1.00\linewidth]{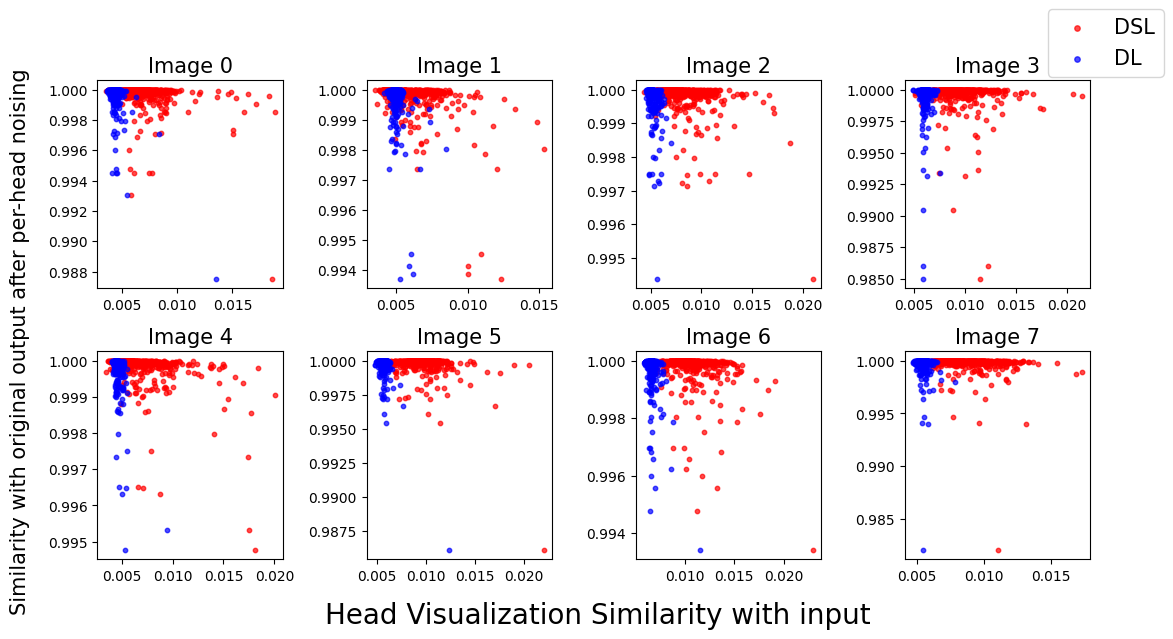}
    \caption{\textit{Correlation between the head ablation effect and the head visualization (flower overlay).}}
    \label{fig:eval1}
    \vspace{-5mm}
\end{figure}

Figure~\ref{fig:eval1} shows that ablating attention heads whose visualizations closely resemble the input image (red dots, DSL) result in a larger loss in the cosine similarity with the original output. However, not every head with high similarity yields a significant loss in similarity in the output space; this variability likely reflects the Hydra effect~\cite{mcgrath2023hydraeffectemergentselfrepair}. In contrast, Diffusion Lens (blue dots) produces a more concentrated distribution, showing less correlation between head visualizations and output similarity. This result aligns with the observation in Section~\ref{sec:visualization}.

\subsubsection{Evaluation 2}
\begin{figure}
    \centering
    \includegraphics[width=1.00\linewidth]{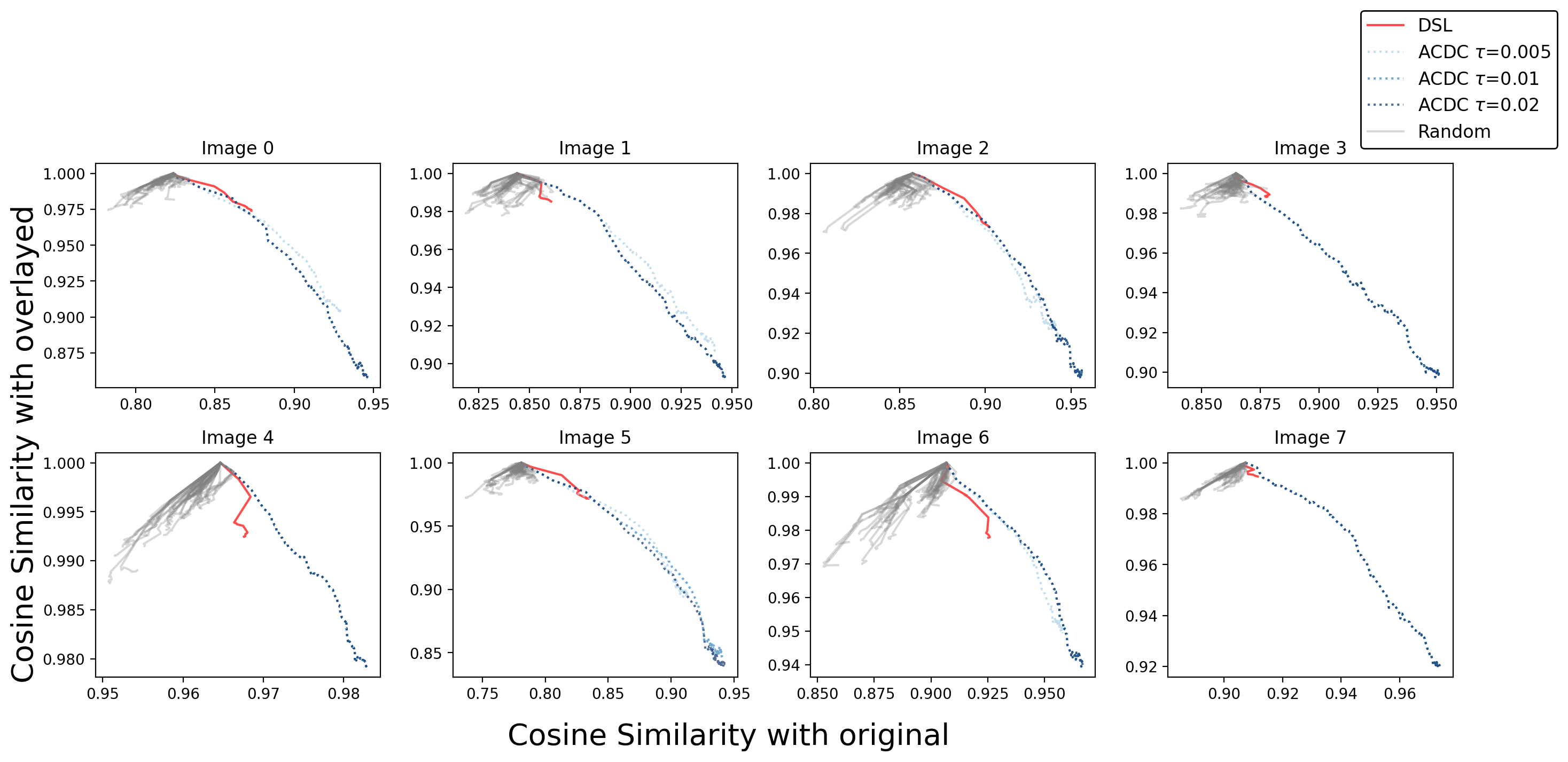}
    \caption{\textit{Comparison of the ViT output trajectories obtained from DSL, ACDC-like, and random ablation strategies for removing overlay information from flower-overlayed images.} DSL consistently steers the output closer to the original image compared to random ablation, closely approaching the performance of the ACDC-like intervention.}
    \label{fig:eval2}
    \vspace{-6mm}
\end{figure}

Figure~\ref{fig:eval2} shows how the ViT's output moves from the overlayed image back toward the original ImageNet image as we ablate heads successively for a flower overlay (similar results are observed for other overlays; see Supplementary). Ablating attention heads whose visualizations resemble the overlay effectively removes the overlay’s influence, shifting the output closer to the original image. The DSL ablation trajectories nearly overlap the optimal case (ACDC-like), consistently stay above the random baseline, despite the difficulties of visualizing early-layer contributions.

\section{Conclusions and Discussion\label{sec:discussion_and_conclusion}}
We introduced Diffusion Steering Lens (DSL) to visualize direct submodule contributions in ViTs. Unlike Diffusion Lens, DSL reliably captures individual effects via representation steering and submodule patching. 

\textbf{Limitations}. The steering operation may inadvertently alter representation content, and the diffusion-based decoder can introduce artifacts. Future work will address these issues, explore the visualization of image tokens for understanding early-layer processing, validate consistency with text-based methods~\cite{Gandelsman2023-gt, Gandelsman2024-rz}, and extend DSL to models without pretrained diffusion decoders.

\section*{Acknowledgments}
This work was supported by JST, Moonshot R\&D Grant Number JPMJMS2012. We also thank the ViT-Prisma Library~\cite{joseph2023vit} contributors for their support.



{
    \small
    \bibliographystyle{ieeenat_fullname}
    \bibliography{main}
}

\clearpage
\setcounter{page}{1}
\maketitlesupplementary

\begin{figure}
    \vspace{2mm}
    \includegraphics[width=1.0\linewidth]{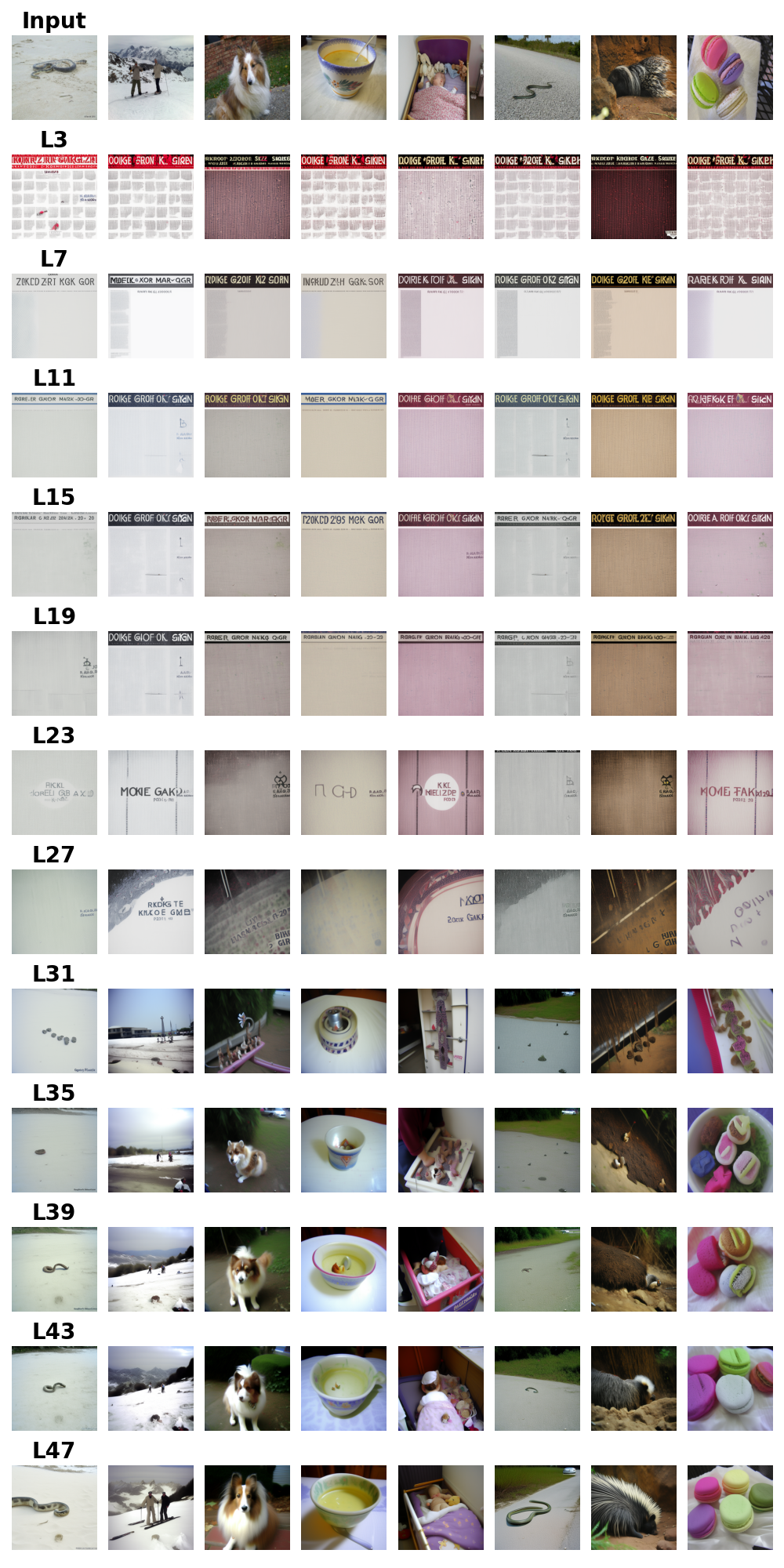}
    \caption{\textit{Resid post visualizations using Diffusion Lens~\cite{Toker2024-ve} (every 4 layer)}}
    \label{fig:diffusion_lens_resid_post_every_4_layers}
    
\end{figure}

\begin{figure}
    \includegraphics[width=1.0\linewidth]{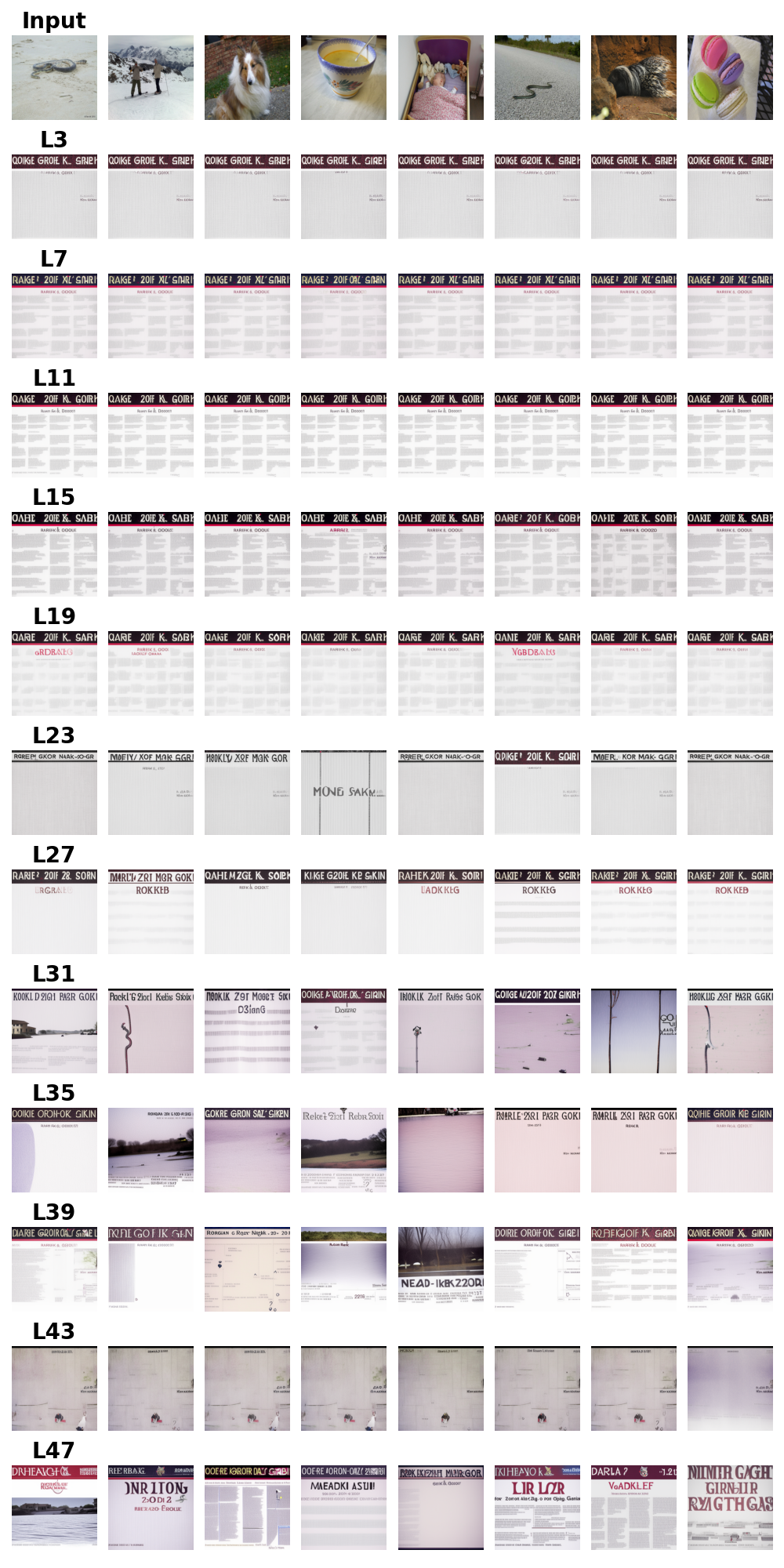}
    \caption{\textit{MLP out visualizations using Diffusion Lens~\cite{Toker2024-ve} (every 4 layer)}}
    \label{fig:diffusion_lens_mlp_every_4_layers}
    \vspace{24.5mm}
\end{figure}

\begin{figure*}
    \centering
    \includegraphics[width=0.9\linewidth]{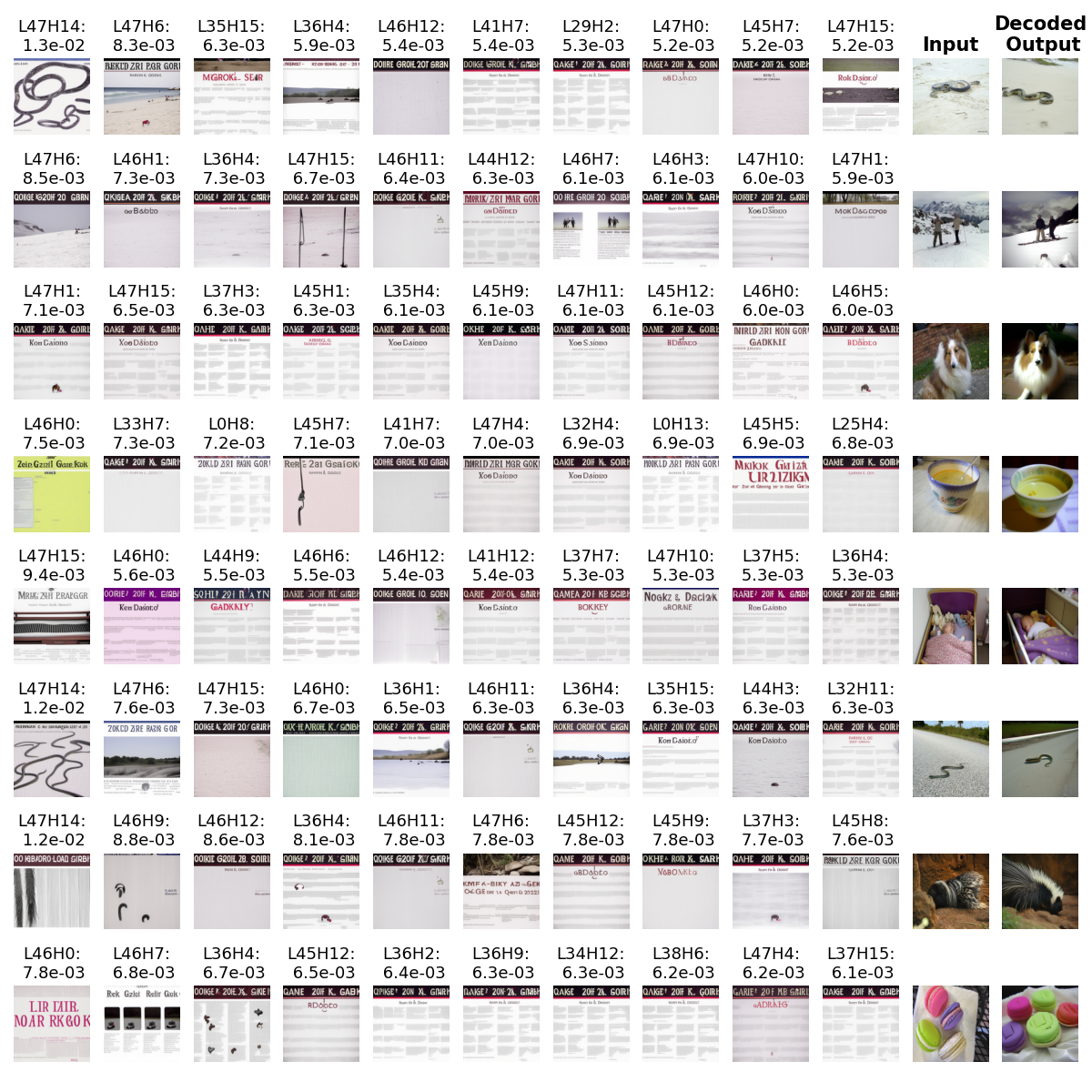}
    \caption{\textit{Top 10 heads in similarity with input when using Diffusion Lens.}}
\end{figure*}

\begin{figure*}
    \centering
    \includegraphics[width=0.9\linewidth]{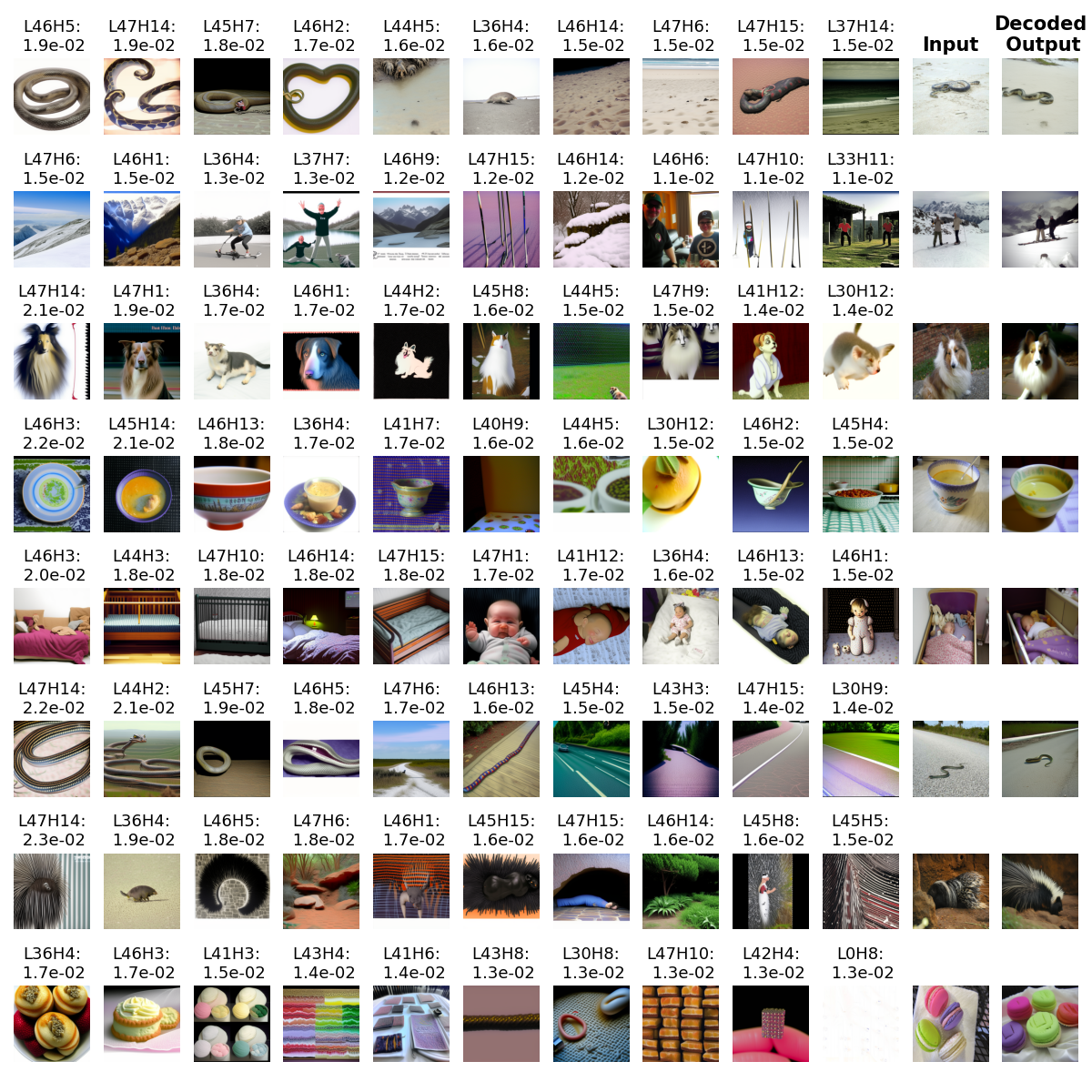}
    \caption{\textit{Top 10 heads in similarity with input when using Diffusion Steering Lens.}}
\end{figure*}

\begin{figure*}
    \centering
    \includegraphics[width=0.7\linewidth]{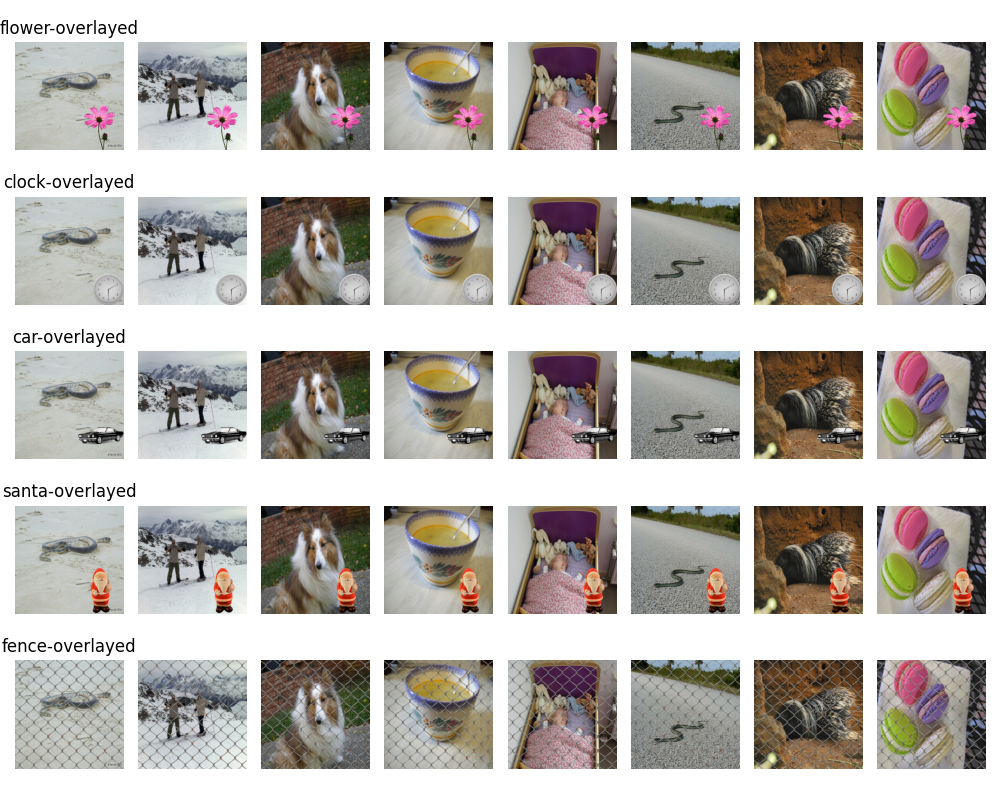}
    \caption{\textit{Images with overlays.}}
\end{figure*}

\begin{figure*}
    \centering
    \includegraphics[width=0.7\linewidth]{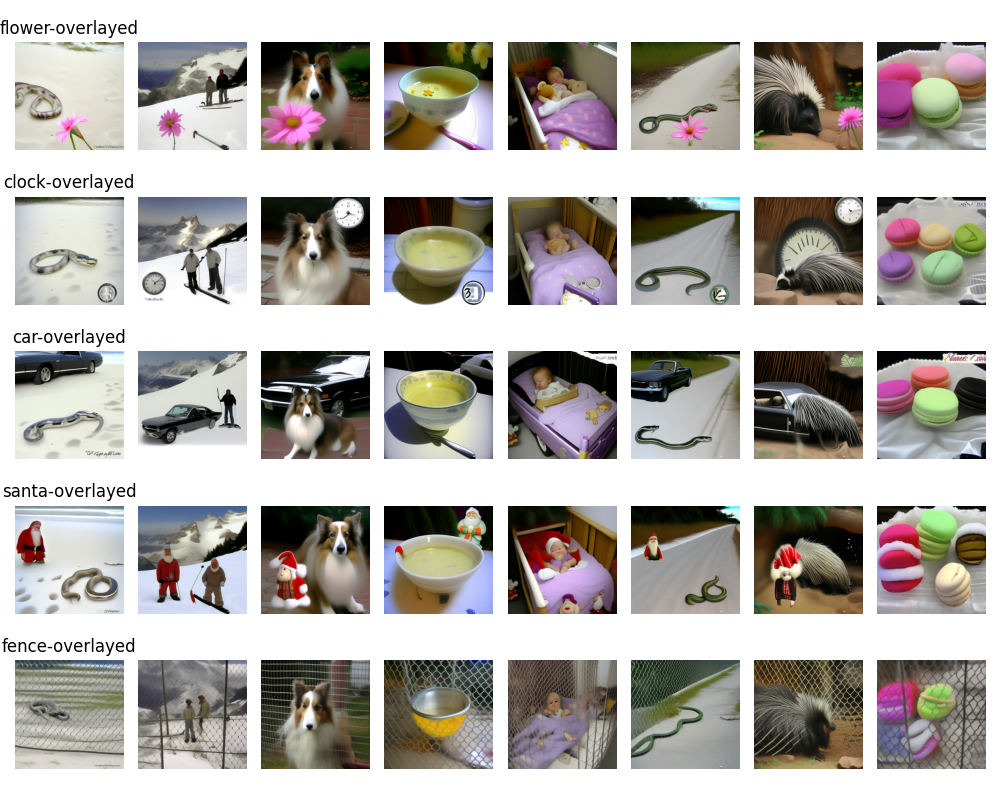}
    \caption{\textit{Reconstruction results of images with overlays.}}
\end{figure*}

\begin{figure*}
    \centering
    \includegraphics[width=1\linewidth]{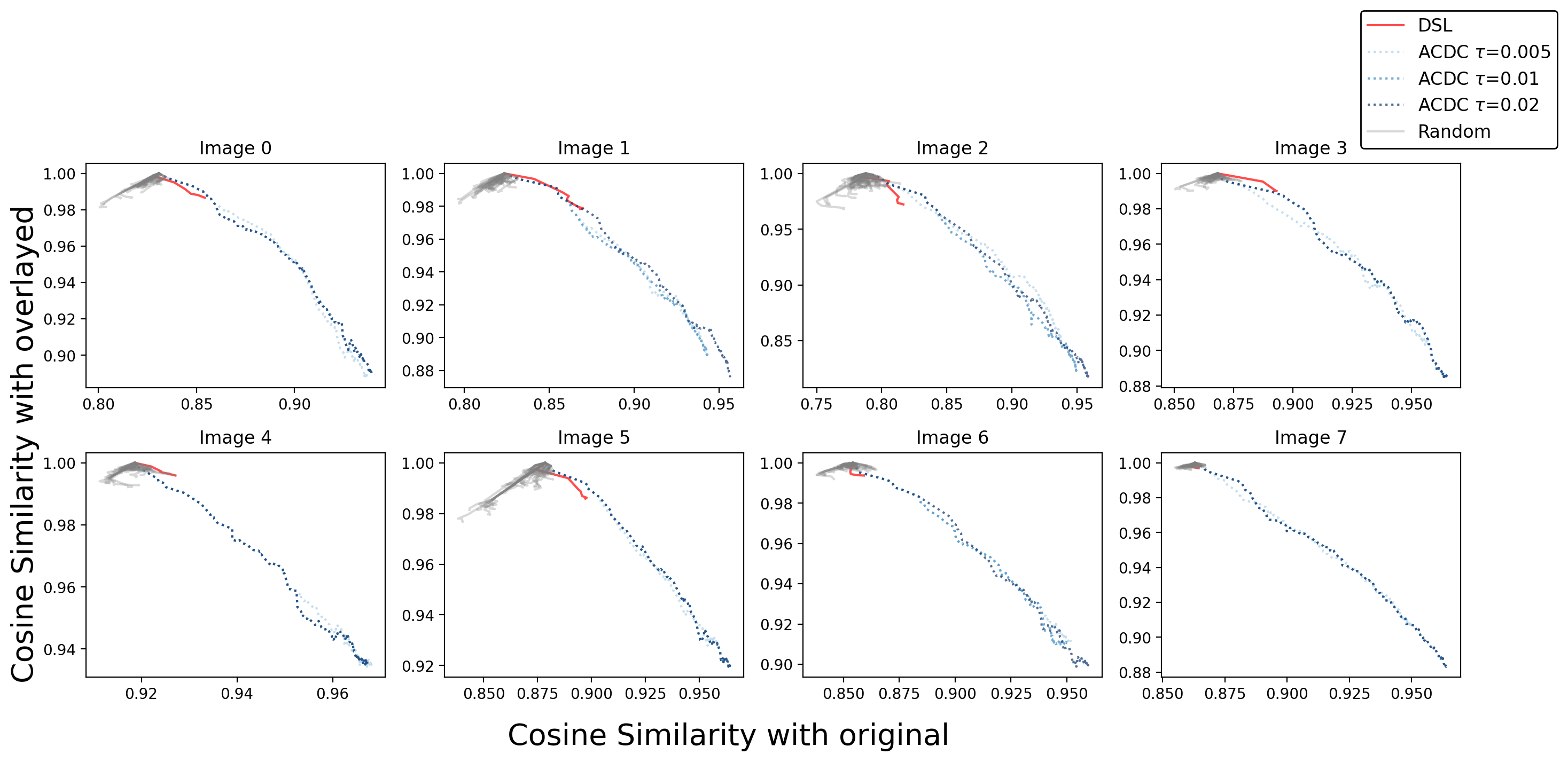}
    \caption{\textit{Comparison of DSL, ACDC-like, and random ablation strategies for removing overlay information from clock-overlayed images.}}
\end{figure*}

\begin{figure*}
    \centering
    \includegraphics[width=1\linewidth]{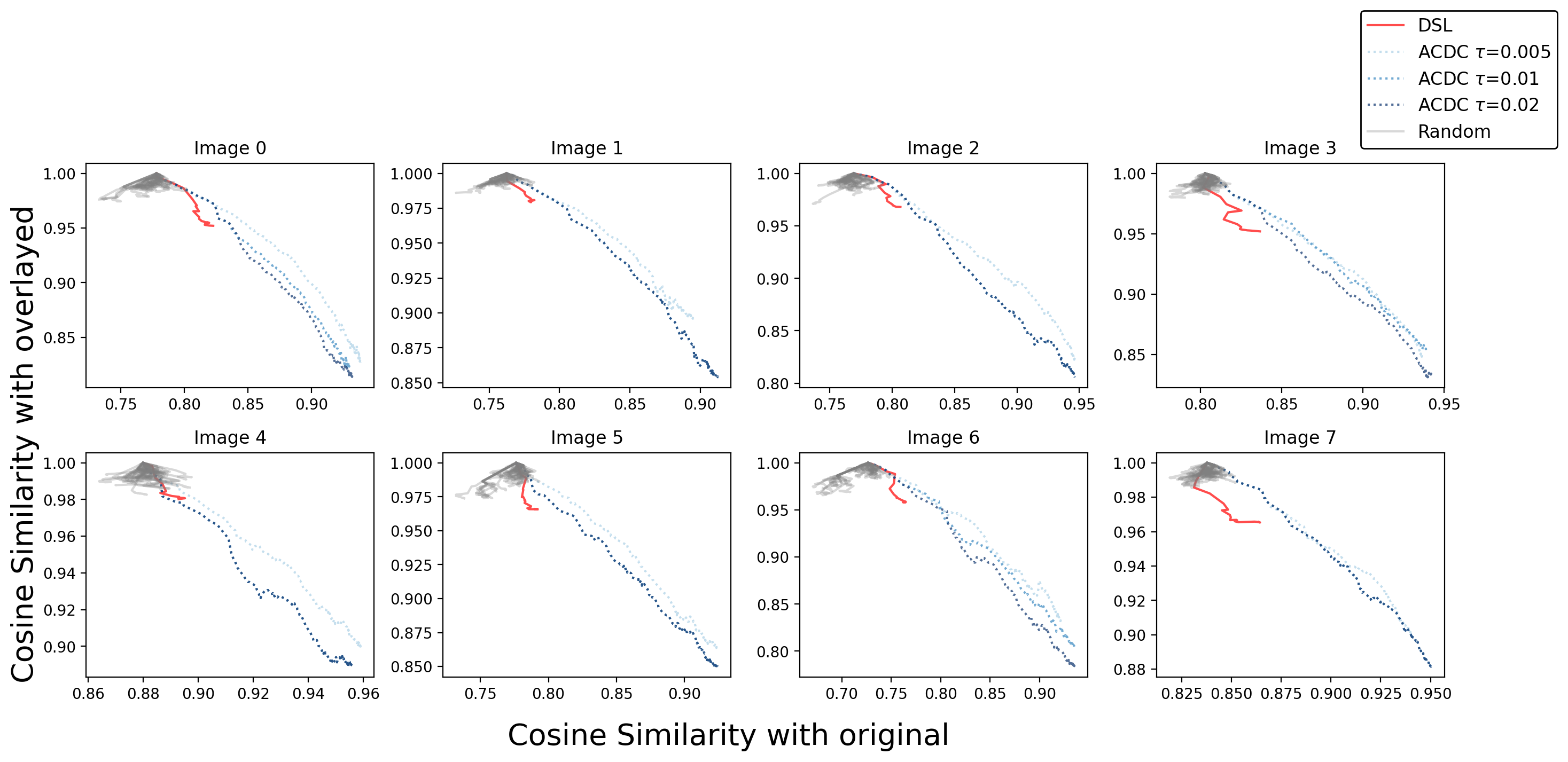}
    \caption{\textit{Comparison of DSL, ACDC-like, and random ablation strategies for removing overlay information from car-overlayed images.}}
\end{figure*}

\begin{figure*}
    \centering
    \includegraphics[width=1\linewidth]{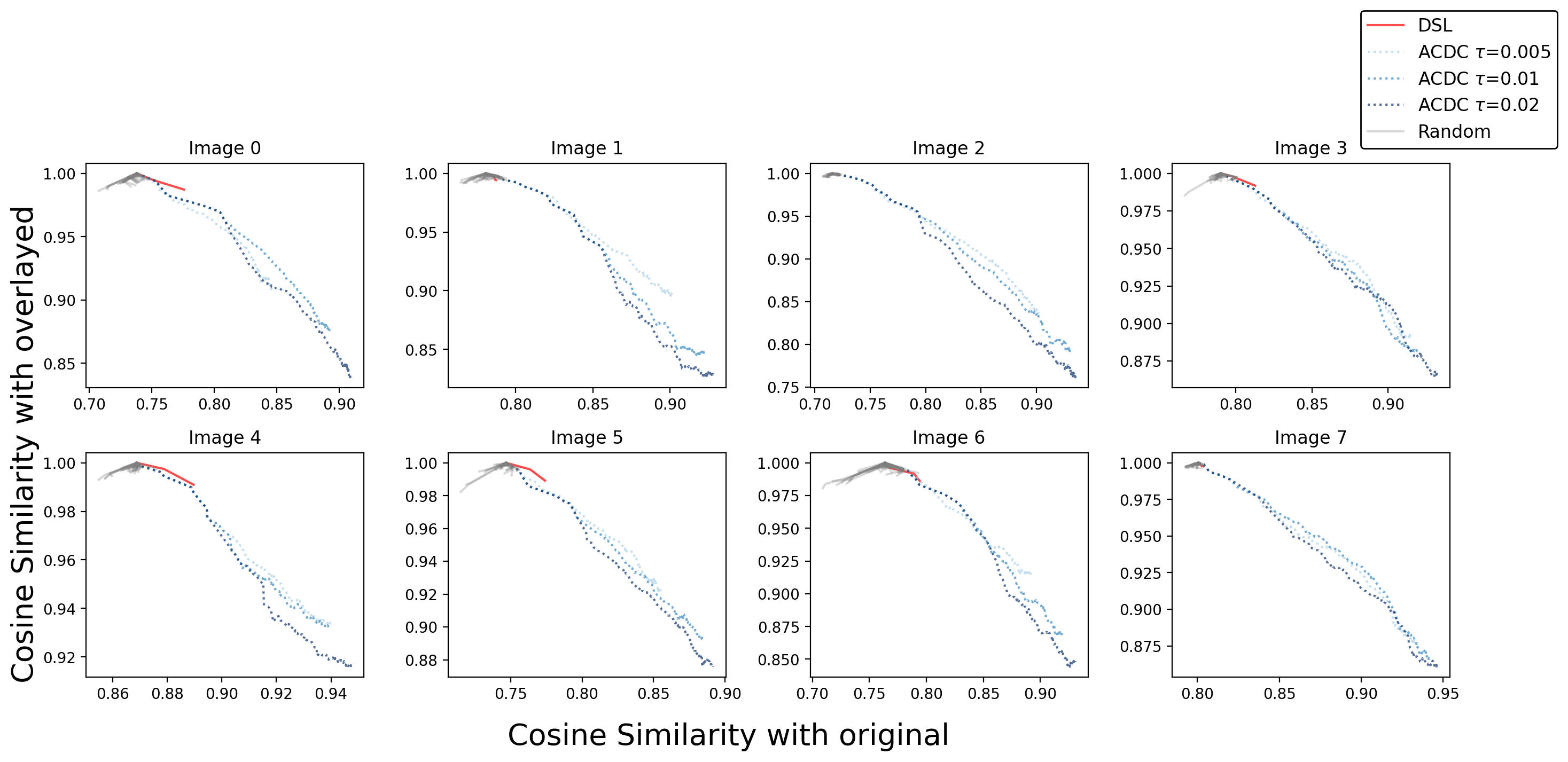}
    \caption{\textit{Comparison of DSL, ACDC-like, and random ablation strategies for removing overlay information from santa-overlayed images.}}
\end{figure*}

\begin{figure*}
    \centering
    \includegraphics[width=1\linewidth]{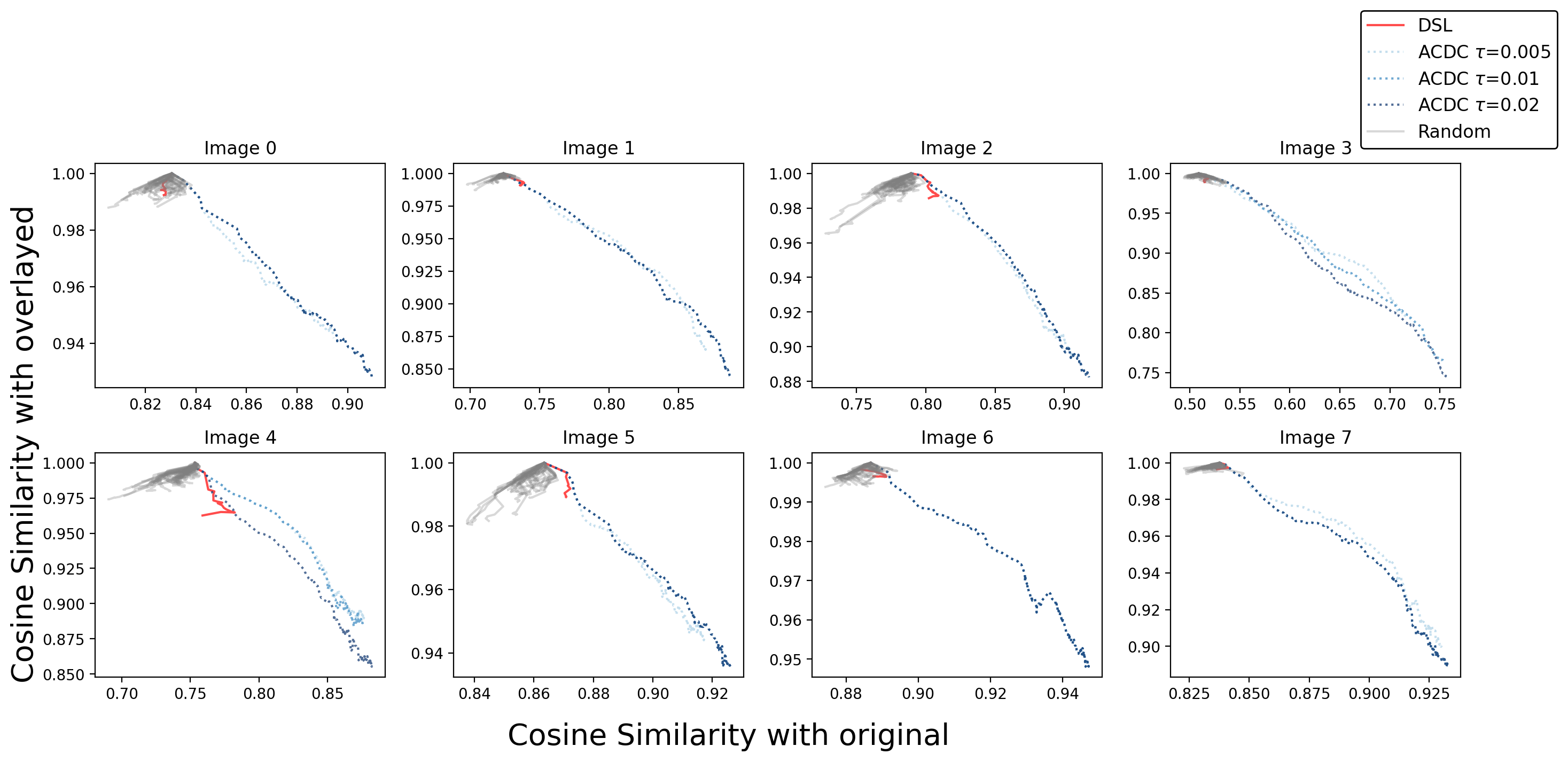}
    \caption{\textit{Comparison of DSL, ACDC-like, and random ablation strategies for removing overlay information from fence-overlayed images.}}
\end{figure*}

\end{document}